\documentclass{article}
\usepackage{geometry}
\usepackage{url}

\newcommand{\projecturl}{\url{https://github.com/bermanmaxim/superpixPool}}

\title{Efficient semantic image segmentation with superpixel pooling}

\author{Mathijs Schuurmans\quad Maxim Berman\quad Matthew B.\ Blaschko\\
 Dept. ESAT, Center for Processing Speech and Images\\
 KU Leuven, Belgium\\
\texttt{mathijs.schuurmans@student.kuleuven.be}\\
\texttt{\{maxim.berman, matthew.blaschko\}@esat.kuleuven.be}%
}

\usepackage{cleveref} 
\usepackage{booktabs}
\usepackage{multirow}
\newcommand{\ve}[1]{\mathbf{#1}}
\usepackage{algorithm}
\usepackage{algpseudocode}
\usepackage{xargs}
\usepackage{graphicx}
\usepackage{subfigure}
\usepackage{amsmath}
\usepackage{dsfont}
\usepackage{lipsum}
\usepackage{microtype}
\usepackage{siunitx}

\usepackage{etoolbox}
\AtBeginEnvironment{tabular}{\small}

\usepackage[colorinlistoftodos,prependcaption,textsize=tiny]{todonotes}
\newcommandx{\info}[2][1=]{\todo[linecolor=OliveGreen,backgroundcolor=OliveGreen!25,bordercolor=OliveGreen,#1]{#2}}

\newcommandx{\overview}[2][1=Section overview]
{\todo[linecolor=Blue,backgroundcolor=Blue!25,bordercolor=Blue,inline,caption={#1}]{#2}}

\newcommandx{\question}[2][1=]
{\todo[linecolor=Red,backgroundcolor=Red!25,bordercolor=Red,#1]{#2}}

\begin{document}

\maketitle

\begin{abstract}
In this work, we evaluate the use of superpixel pooling layers in deep network architectures for semantic segmentation. Superpixel pooling is a flexible and efficient replacement for other pooling strategies that incorporates spatial prior information. 
We propose a simple and efficient GPU-implementation of the layer and explore several designs for the integration of the layer into existing network architectures. We provide experimental results on the IBSR and Cityscapes dataset, demonstrating that superpixel pooling can be leveraged to consistently increase network accuracy with minimal computational overhead.
Source code is available at \projecturl.
\end{abstract}

\section{Introduction}

Superpixels have been a popular method of incorporating spatial priors in a wide variety of computer vision problems.  This has been applied both to influence the resulting statistical models to favor spatial smoothness, but also for computation reasons.  For graph cuts based inference, for example, superpixels have been widely employed to reduce the cubic time worst case computational cost.  They have somewhat fallen out of favor with the recent popularization of deep neural networks, for classification, segmentation, and many other vision tasks.
However, the statistical and computational advantages of superpixels can also apply in the context of deep networks.

A key reason for this is that deep neural networks designed for computer vision tasks have grown increasingly deep and complex over the last years. Although this allows for an increase in accuracy on a variety of problems, it typically also comes at an increased computational cost. Our aim is therefore to improve the accuracy of networks by simplifying the segmentation problem rather than building more complex models. We propose to do this by means of a superpixel pooling layer. Such a layer has two main benefits: information grouping and introduction of a prior on the segmented output. This prior favors segmentations that preserve the edges provided by the superpixels.
Information grouping and edge preservation are, however, conflicting objectives, since larger superpixels group more information but will generally not adhere to boundaries as well as smaller superpixels. In our experiments, we attempt to gain insight into this trade-off and see how it affects segmentation accuracy for different types of networks. 

We generally use the term~\emph{superpixel} for 2D image segmentation, and~\emph{supervoxel} for  volumetric image segmentation. 
We start by describing the superpixel pooling layer and propose a simple but efficient GPU-implementation for the forward and backward pass. We then propose several ways of integrating the layer into existing CNN architectures. Finally, we empirically evaluate these modifications using the \textit{VoxResNet}~\cite{chen2016voxresnet} and ENet~\cite{paszke2016enet} architectures.

\subsection{Related work}

Over the years, many different superpixel segmentation algorithms have been developed. A recent benchmark study~\cite{stutz2017superpixels} classifies 24 of the state-of-the-art methods by their high-level approach and constructs an overall ranking of the algorithms. The overall top-performing candidates from this benchmark are \textit{Efficient Topology Preserving Segmentation} (ETPS)~\cite{yao2015real} and \textit{Superpixels extracted via energy-driven sampling} (SEEDS) \cite{van2012seeds}. These are closely followed by perhaps the most popular superpixel algorithm called \textit{Simple Linear Iterative Clustering} (SLIC)~\cite{achanta2012slic}. Aside from its overall segmentation accuracy that is competitive with SEEDS and ETPS, SLIC has several additional benefits for our purposes. These include its simplicity, native support for volumetric images and its parallelizability. The latter has led to gSLICr~\cite{ren2015gslicr}, a GPU-implementation of the SLIC algorithm. Such an implementation allows to provide superpixels to the network with at a minimal time cost. 
The first network architecture we use in our experiments below is called \textit{VoxResNet}~\cite{chen2016voxresnet}. This is a residual neural network~\cite{he2016deep} that was designed to work directly on volumetric images and reaches state-of-the-art performance in segmentation of anatomical regions in MRI brain scans. We use this network along with two simplified variants of it to test the efficacy of the supervoxel pooling layer on volumetric image segmentation. 
The second network architecture, called \textit{ENet}~\cite{paszke2016enet} uses a encoder-decoder structure and was designed for the purpose of real-time segmentation. This design objective makes it a suitable test case as it aligns well with the purpose of the superpixel pooling layer.

Recently, the idea of using superpixel-wise pooling has also been explored in different contexts, such as semi-supervised segmentation~\cite{kwak2017weakly} and image classification~\cite{yang2017improved}. We propose to apply the technique directly in a fully supervised semantic segmentation setting. Moreover, we provide a full description of the layer, including implementation details and extensive empirical evaluation of the method on both 2D and 3D images. This includes specific designs for the integration of the superpixel pooling layer into existing CNN architectures. 

\section{Superpixel pooling} \label{sec:pooling}

\begin{figure}[ht]
\centering
\includegraphics[width=0.7\textwidth]{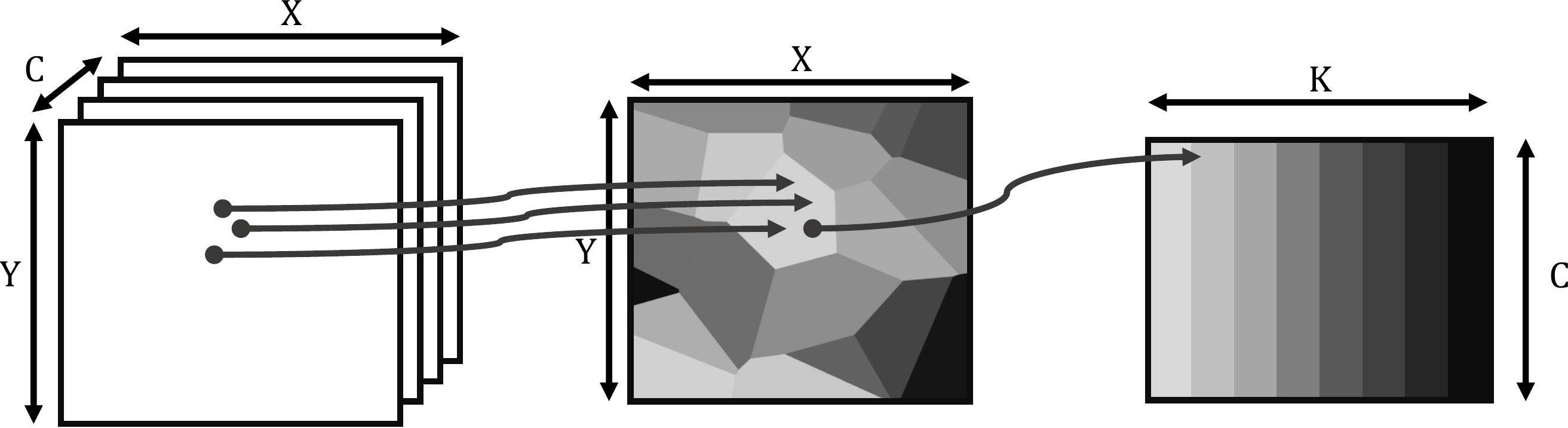}
\caption{Superpixel pooling layer on a 2D image. The arrows denote the flow of information of a few pixels in the first channel to the output in the forward pass through the layer.}
\label{fig:supvoxpool}
\end{figure}
The general idea of superpixel pooling is very similar to regular pooling. The features over a local region (i.e. a superpixel) are aggregated by means of a reduction function (e.g. $\max(\ve{\cdot})$ or $\mathrm{avg}(\ve{\cdot})$). This step reduces the information in the image from one feature vector per pixel to one feature vector per superpixel.
Consider the input image $\ve{I} \in \mathds{R}^{C \times P}$,
composed of $C$ channels and $P$ pixels. 
A superpixel segmentation is represented by a one-channel image  
$\ve{S} \in L^{P}$ 
where $L = [1, K]$ are the integer labels of each superpixels, such that pixel $i$ belongs to superpixel $k$ if $S_i = k$. 
Given these inputs, the superpixel pooling layer returns an array $\ve{P} \in \mathds{R}^{C \times K}$, with 
\begin{align} \label{eq:supvoxpool}
	P_{c,k} = \mathit{reduce} \,\{ I_{c, i} \,|\,i: S_i = k\}
\end{align}
where $\mathit{reduce}$ is a reduction function such as $\mathit{average}$ or $\mathit{max}$.
Put simply, the value at the $c$th row and $k$th column of the output array is the output of the reduction function applied to the values at channel $c$ of the pixels in superpixel $k$. This principle is illustrated in~\Cref{fig:supvoxpool}.

We detail the computation of the gradient for the back-propagation through the superpixel pooling layer with max pooling and with average pooling.
\begin{description}
	\item[max pooling] Denoting $i^*$ the pixel for which $I_{c,i^*} = \mathit{max} \,\{ I_{c, i} \,|\,i: S_i = k\}$, 
    we have $P_{c,k}=I_{c,i^*}$, therefore the backward gradient is given by
    \begin{align} \label{eq:bw_max}
        \frac{\delta\,P_{c,k}}{\delta I_{c',i}} = \begin{cases}
        					  1 & \textrm{ if } i = i^* \text{ and } c' = c \\
                              0			   & \textrm{ otherwise. }
        				   \end{cases}
    \end{align}
    \item[average pooling] In this case we can write
    $
    	P_{c,k} = \frac{1}{N(k)} \sum_{i:\, S_i = k} I_{c, i},
     $
     where $N(k) = |\{i:\, S_i = k\}|$.
     Therefore the computation of the derivative for the backwards pass reduces to
     \begin{align} \label{eq:bw_avg}
     	\frac{\delta\,P_{c,k}}{\delta I_{c',i}} = \begin{cases}
        					  1/N(S_i) & \text{if $S_i = k$ and $c' = c$}, \\
                              0			   & \textrm{ otherwise. }
        				   \end{cases}
     \end{align}
\end{description}

\paragraph{CPU implementation}

We experiment with two CPU--implementations for the superpixel pooling layer. The first implementation is based on the \texttt{ndimage} package from the Scipy library~\cite{scipy} for Python.
This results in a very straightforward implementation, which is useful for the verification of the more highly optimized versions of the layer.
The second version of the layer is implemented in Python and accelerated by the use of the Numba compiler~\cite{numba}. The execution time is linear in the number of pixels below the pooling layer.
 The maximizing pixels $i^*$ during max-pooling, or the pixel counts $N(k)$ during average-pooling, are computed and stored during the forward pass. In the backward pass, they are reused to evaluate the gradients according to Equation~\eqref{eq:bw_max} or \eqref{eq:bw_avg} respectively. 

\begin{figure}[ht]
\centering
\includegraphics[width=0.8\textwidth]{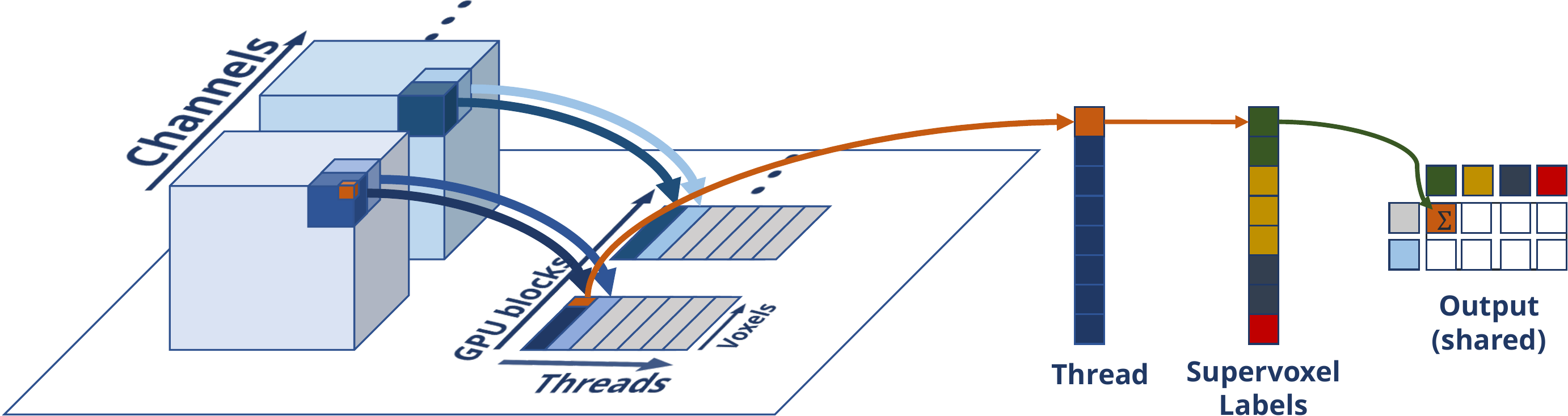}
\caption{GPU implementation of a supervoxel pooling layer. GPU blocks treat different channels of the input volume, while threads within a block are assigned an image cube along a regular grid. 
Each thread goes over the pixels of the cube assigned to it (in blue) and the associated supervoxel assignments, updating the shared copy of the running counts/maximum within the superpixel (shared output).}
\label{fig:GPU_scheme}
\end{figure}

\paragraph{GPU implementation}
Additionally, we propose a parallelized GPU-implementation, which is crucial for efficient training and inference of a CNN with a superpixel pooling layer.
In the forward pass, our implementation is designed to take advantage of parallelism in the following ways.
(1)~Since the outputs of the different channels are mutually independent, they do not need to share any memory. They can thus be processed by separate GPU blocks. 
(2)~The image is divided along a regular grid, and each thread of the GPU block is assigned a cell of this grid to process. 
Threads within a block share a synchronized read-write array of size $K$ keeping track of the counts (for average) or the maximum (for max-pooling) for each of the superpixels.
This process is illustrated in the case of a volumetric image (supervoxel pooling) in~\Cref{fig:GPU_scheme}.

To avoid racing conditions when writing on the output array, we employ \textit{atomic functions}. These functions temporarily lock the memory address while a thread is performing a writing operation on it. Generally speaking, atomic functions can hurt performance since they enforce some amount of serialization. This is mainly an issue when many threads have to access a the same address.  
This is not the case here: since superpixels are spatially concentrated, only a limited number of threads will update the entries relevant to each superpixel. 

As in the CPU-implementations, the necessary information for the backward pass is cached during the forward pass and reused during backpropagation. The gradients of all combinations of pixels and channels can be computed in parallel on the GPU.

\section{Experiments}

\subsection{Efficiency of the superpixel pooling layer}
\begin{figure}[ht]
\centering
\subfigure[Runtime vs. image size]{
\includegraphics[height=2.4cm]{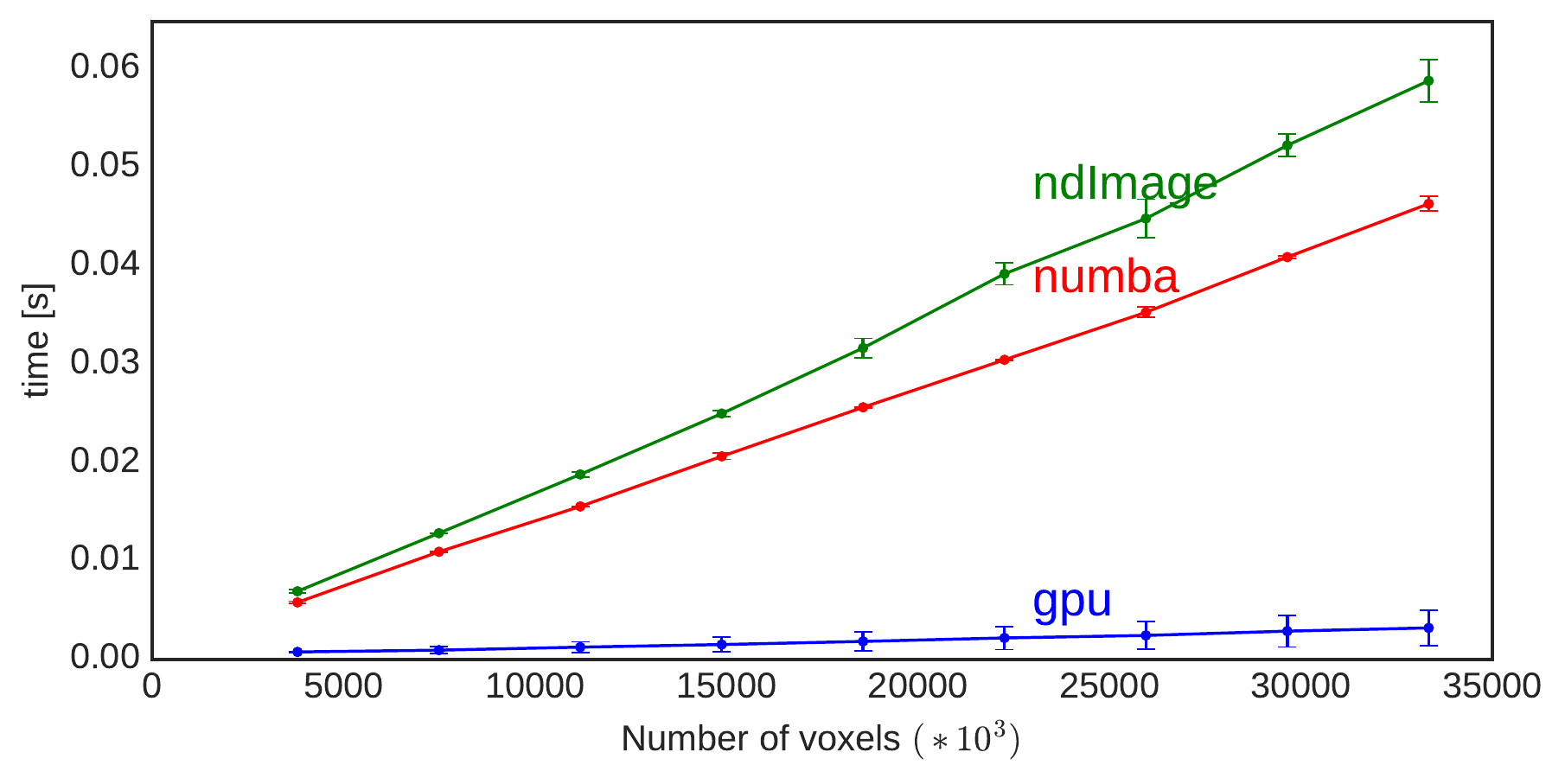}
}
\label{fig:GPUN}
\subfigure[Runtime vs. number of clusters]{\includegraphics[height=2.4cm]{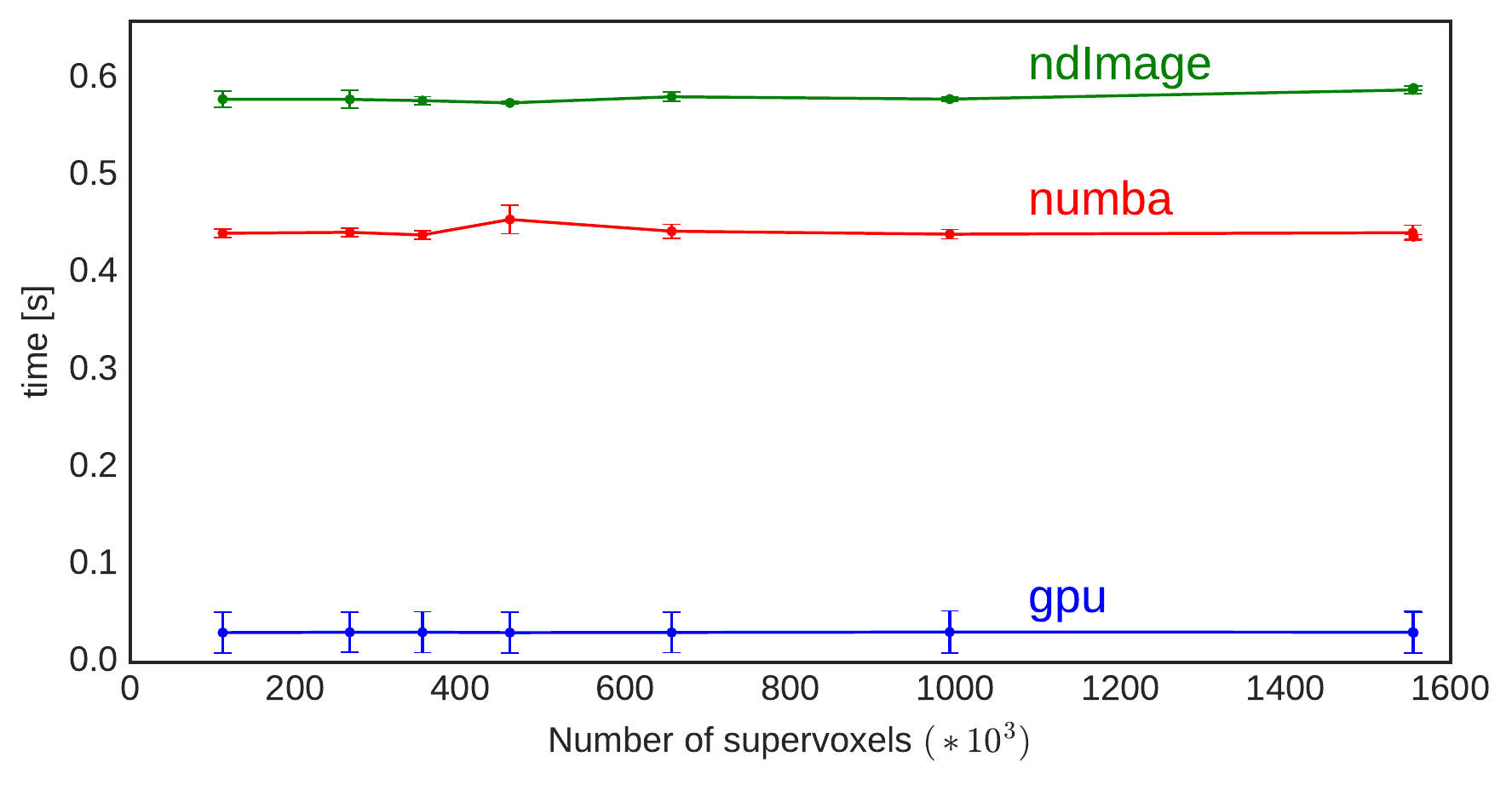}}
\label{fig:GPUK}
\caption{Runtimes for different implementations of the superpixel pooling layer.}
\label{fig:GPUTime}
\end{figure}

\Cref{fig:GPUTime} shows the processing time of our implementations of the pooling layer as a function of the number of pixels $P$ and number of clusters $K$. The GPU implementation vastly outperforms the CPU implementations, yielding a 16-fold reduction in processing time.
All three methods scale linearly with $P$ and are independent of $K$ for all values considered in practice. 
\subsection{Integration of SupVoxPool with VoxResNet} \label{sec:intVRN}
\subsubsection{Experimental set-up}

\begin{figure}[ht]
\centering
\includegraphics[width=0.6\textwidth]{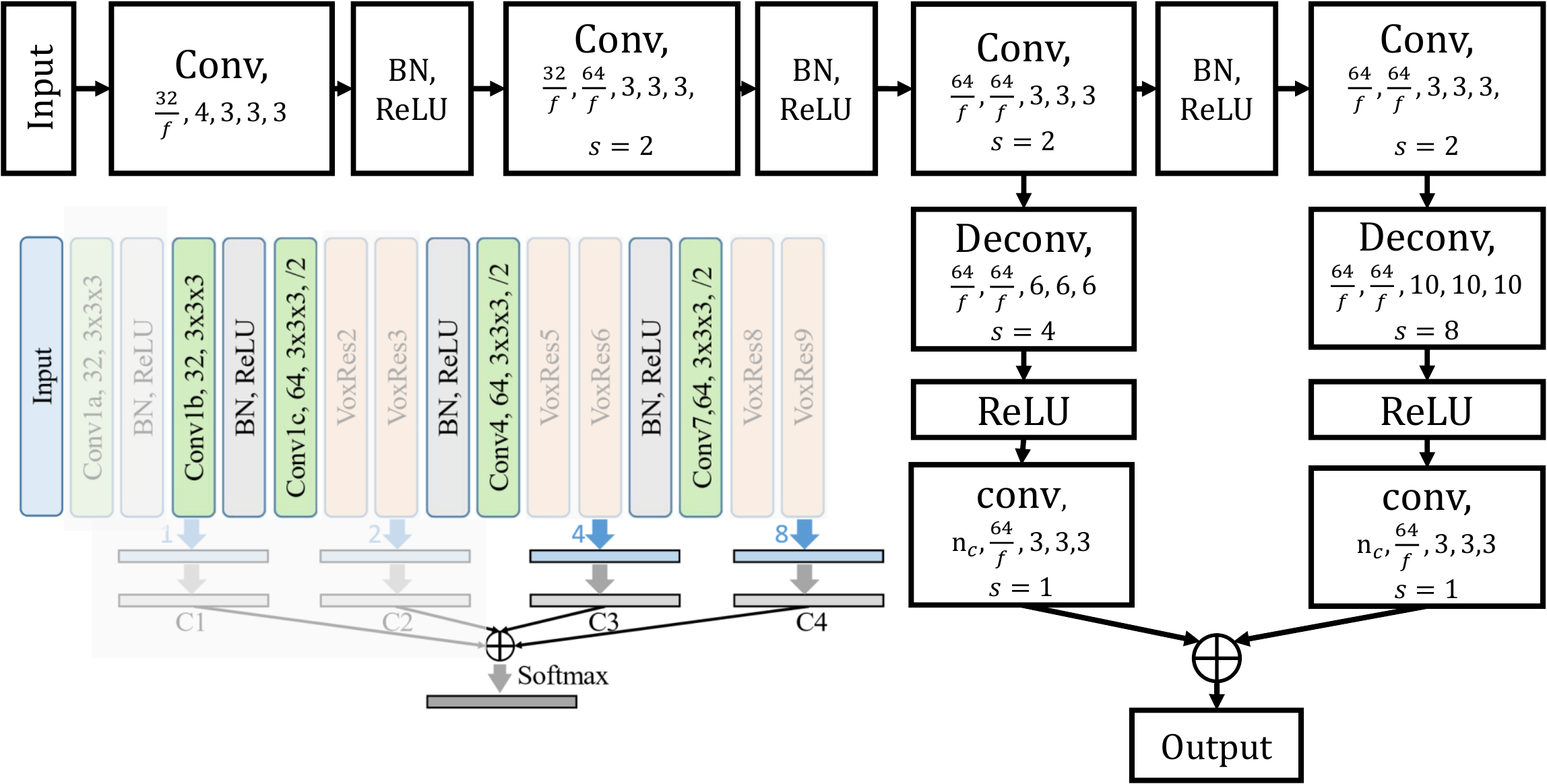}
\caption{Detailed architecture of the reduced \textit{VoxResNet}~\cite{chen2016voxresnet}. The factor $f$ can be used to further simplify the network by reducing the dimensionality of the feature space. At the bottom-left, the original architecture is shown with the omitted layers grayed-out.} 
\label{fig:reducedVRN}
\end{figure}

In the following experiments, we consider three variants of the~\textit{VoxResNet} architecture. The first is the full architecture as described in~\cite{chen2016voxresnet}. The other two are reduced versions of this network (see~\Cref{fig:reducedVRN}). In the reduced network, we remove a large number of layers from the network and reduce the dimensionality of the feature space by a factor $1/f$. We denote these reduced variants of the network by ``Reduced ($/f$)''. Doing this allows us to test the efficiency of adding a superpixel pooling layer in contrast to increasing the network complexity.

\begin{figure}[ht]
\centering
\includegraphics[width=0.8\textwidth]{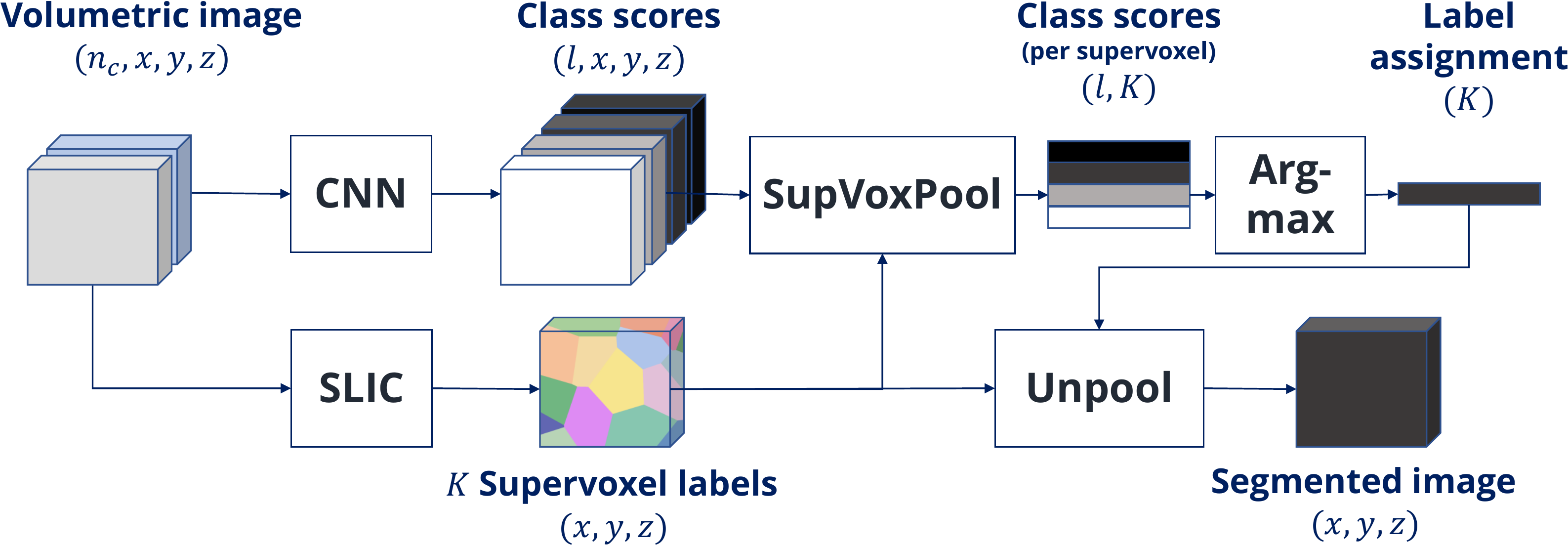}
\caption{Illustration of the segmentation pipeline using the SupVoxPool layer as a postprocessing step (\textit{``Version 1''}).}
\label{fig:postprocess}
\end{figure}
For each of the three basis networks, we consider three ways of integrating the supervoxel pooling (SupVoxPool) layer into the network, denoted as Version 1, 2 and 3. 

\textbf{Version 1} is illustrated in~\Cref{fig:postprocess}. In this scheme, supervoxel pooling is used as a postprocessing step. The network itself remains unchanged. In this scenario, finetuning will only allow the network to update its weights to provide a better input to the SupVoxPool layer.

\begin{figure}[ht]
\centering
\includegraphics[width=\textwidth, trim={0 0.25cm 0 0.25cm},clip]{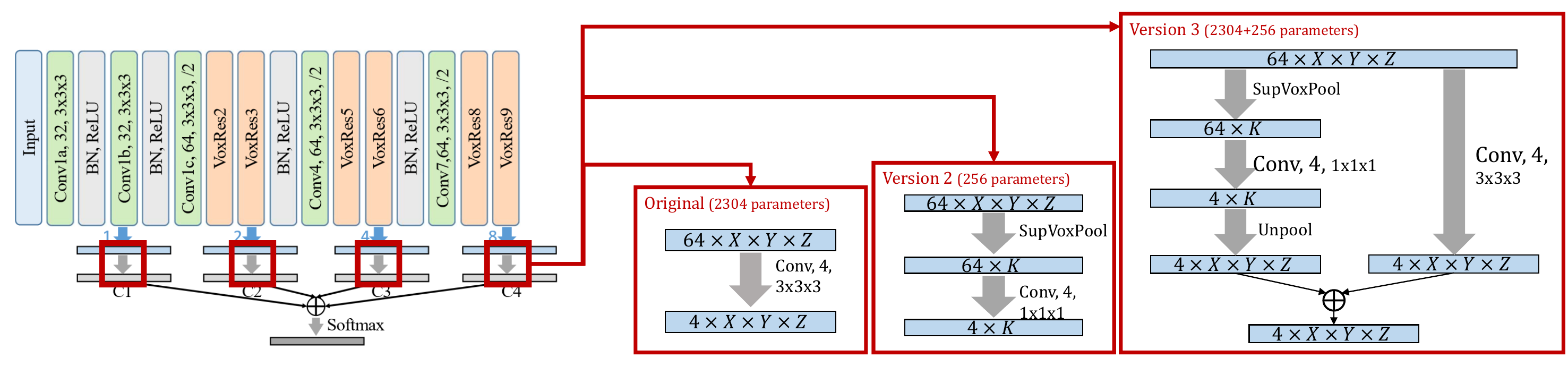}
\caption{Different designs for the integration of the supervoxel pooling layer into the \textit{VoxResNet}~\cite{chen2016voxresnet} architecture.}
\label{fig:integrationVRN}
\end{figure}

In \textbf{Version 2} (Fig.~\labelcref{fig:integrationVRN}), the final $3 \times 3 \times 3$ convolution is replaced with a SupVoxPool layer followed by a fully connected layer. This allows the network to learn a mapping from supervoxel-wise features to class labels.

\textbf{Version 3} combines both approaches. The final layer is split into two separate branches. One branch operates on superpixels and consists of the same components as V2. The other branch operates at the pixel level and is preserved from the base network. In this setup, the network is allowed to learn pixel-wise corrections to the superpixel labels.

The baseline networks (without supervoxels) are all trained from scratch on the Internet Brain Segmentation Repository (IBSR)~\cite{rohlfing2012image}, with 4 classes (BG, CSF, WM, GM). The supervoxel networks are then finetuned from these baseline networks. The reported results are Dice coefficients on test data. They are averaged over a 3-fold cross-validation, where 12 volumes were used for training and model selection. The remaining 6 are used for testing. We used max pooling in our results below as it consistently outperformed average pooling during initial experiments. 

\subsubsection{Results}
A comparison between the accuracies of the different network designs is given in~\Cref{tab:blockDice}. The designs in which the superpixels impose a hard constraint, namely V1 and V3 do not consistently improve the results. It's also in these cases that the used superpixels need to have an Oracle performance that exceeds the baseline network accuracy. This puts a stringent lower bound on the number of supervoxels that can be used in conjunction with a given network.
The hybrid design (V3) increases the dice coefficient of all three network complexities. This increase is more substantial for the reduced networks, for which the use of larger supervoxels is possible.

\begin{table}[ht]
\centering
\caption{Dice coefficients (\%) on IBSR for the different network designs outlined in~\Cref{sec:intVRN} and different shapes and sizes of supervoxels.}
\label{tab:blockDice}
\begin{tabular}{rcccccc}
\toprule
                       & \multicolumn{2}{c}{Full Net} & \multicolumn{2}{c}{Reduced (/4)} & \multicolumn{2}{c}{Reduced (/8)} \\ \midrule
                       & BLOCK             & SLIC     & BLOCK               & SLIC       & BLOCK      & SLIC                \\ \cmidrule(l){2-3} \cmidrule(l){4-5} \cmidrule(l){6-7}
Base network accuracy & \multicolumn{2}{c}{88.0}  & \multicolumn{2}{c}{78.9}      & \multicolumn{2}{c}{72.2}      \\
V1                     & 83.4           & 84.7   & 71.8             & 78.3     & 67.9     & 73.1             \\
V2                     & 83.8           & 86.2   & 66.0             & 72.2     & 62.6     & 68.4             \\
V3                     & \textbf{90.2}  & 88.9   & \textbf{82.7}    & 82.3     & 75.2     & \textbf{76.9} \\
\midrule
\textit{Supervox. Oracle perf.} &      \textit{89.0}      & \textit{91.1}  & \textit{79.6}             & \textit{87.8}    & \textit{72.7}    & \textit{87.8}
\\ \bottomrule
\end{tabular}
\end{table}

To compare the relative importance of the geometric prior versus information grouping, the results on superpixels of infinite compactness (i.e. cubes) are added under the column header `BLOCK' in~\Cref{tab:blockDice}. These BLOCK-superpixels essentially disregard the image boundaries and thus only help in terms of information grouping. In the case of V1 and V2, this further harms the performance of the network as it takes away the possibility of following fine contours. In the case of V3, the networks using cubic superpixels can outperform the networks using SLIC superpixels. We observe that this is increasingly true for the more complex networks since they have more flexibility to fill in precise shape information.

\begin{figure}[ht]
\centering
\subfigure[BLOCK (79.57\%)]{\includegraphics[width=0.25\textwidth, trim={0 0.8cm 0 0.8cm},clip]{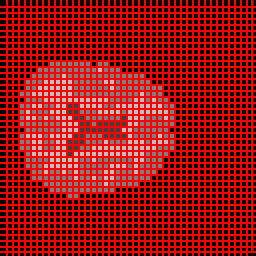}
\label{fig:svxBLOCKMed}}
\subfigure[SLIC (87.76\%)]{\includegraphics[width=0.25\textwidth, trim={0 0.8cm 0 0.8cm},clip]{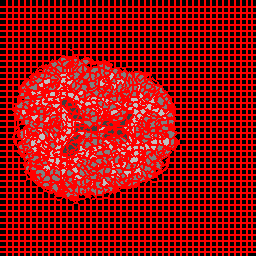}
\label{fig:svxSLICMed}}
\subfigure[SLIC (84.13\%)]{\includegraphics[width=0.25\textwidth, trim={0 0.8cm 0 0.8cm},clip]{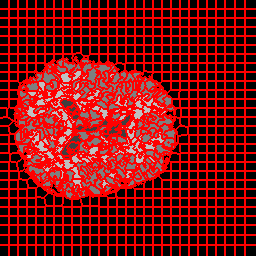}
\label{fig:svxSLICLow}}
\caption{Illustration of the different supervoxel shapes used in the experiment of~\Cref{tab:comparison}.}
\label{fig:svxIllust}
\end{figure}

\begin{table}[ht]
\centering
\caption{Dice coefficients (\%) of the V3 networks using different shapes and sizes of superpixels.}
\label{tab:comparison}
\begin{tabular}{lllll}
\toprule
               & Baseline & \multicolumn{1}{c}{\begin{tabular}[c]{@{}c@{}}BLOCK (V3)\end{tabular}} & \multicolumn{2}{c}{\begin{tabular}[c]{@{}c@{}}SLIC (V3)\end{tabular}} \\ \midrule
Full network   & 88.0  & 89.6                                                                   & 88.97                        & \textbf{89.9}                       \\
Reduction (/4) & 78.9   & 82.7                                                                   & 82.3                        & \textbf{83.3}                       \\
Reduction (/8) & 72.2   & 75.2                                                                   & 76.9                        & \textbf{78.9}                       \\
\midrule
\textit{Supervox. Oracle perf.}   &         & \textit{79.6}                                                                   & \textit{87.8}                        & \textit{84.1}                                \\ 
\bottomrule
\end{tabular}
\end{table}
\Cref{tab:comparison} contains a comparison between the dice scores for three types of supervoxels, illustrated in~\Cref{fig:svxIllust}. The first (\ref{fig:svxBLOCKMed}) are blocks, which only provide information grouping. The second (\ref{fig:svxSLICMed}) are high-detail superpixels. These provide more accurate geometric information but slightly less information grouping than the blocks. The third (\ref{fig:svxSLICLow}) are larger superpixels, which allow for more information grouping, while maintaining better boundary adherence than the blocks. The latter shows superior performance, providing a significant increase in the network accuracy.

\begin{figure}[ht]
\centering
\includegraphics[width=0.7\textwidth, trim={0 0.5cm 0 1cm}, clip]{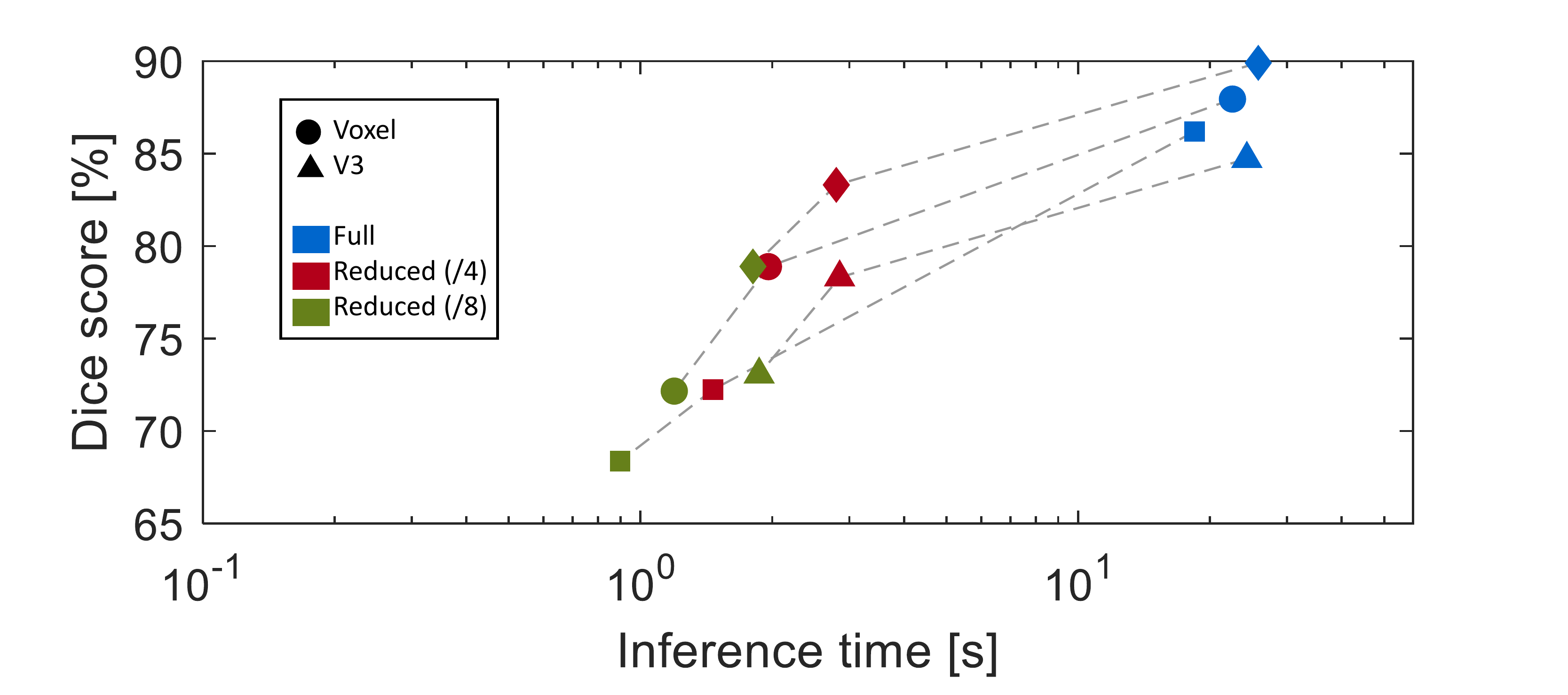}
\caption{Dice coefficients versus inference time of the modified \textit{VoxResNet} architectures.}
\label{fig:diceTimeLowK}
\end{figure}

\Cref{fig:diceTimeLowK} finally illustrates the efficiency of adding the supervoxel pooling layer to smaller networks. The reduced networks have an inference time of 1-3s, compared to ~20-25s for the full network.  

\begin{figure}[ht]
\centering
\includegraphics[width=0.7\textwidth]{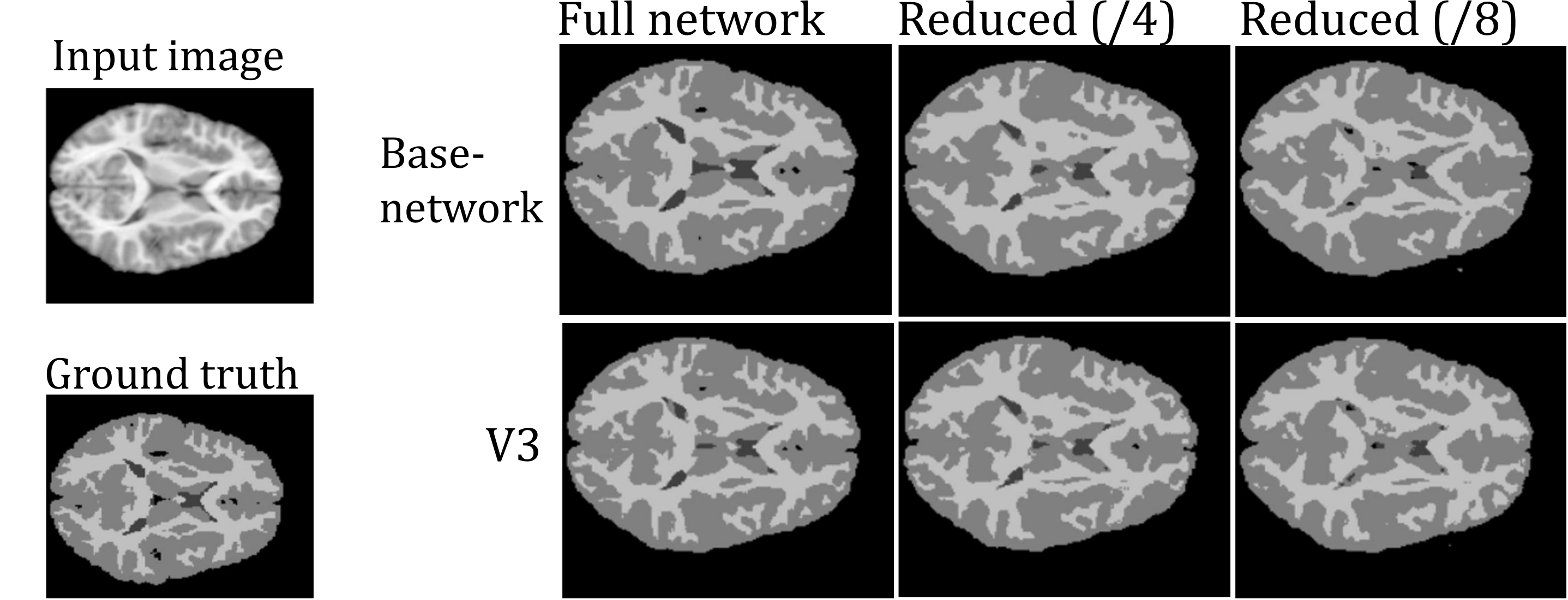}
\caption{Slices of segmentations produced by the different variants of the \textit{VoxResNet} architectures.}
\label{fig:netOuts}
\end{figure}

\subsection{Integration of SupPixPool in ENet}
\begin{figure}[ht]
\centering
\includegraphics[width=0.7\textwidth]{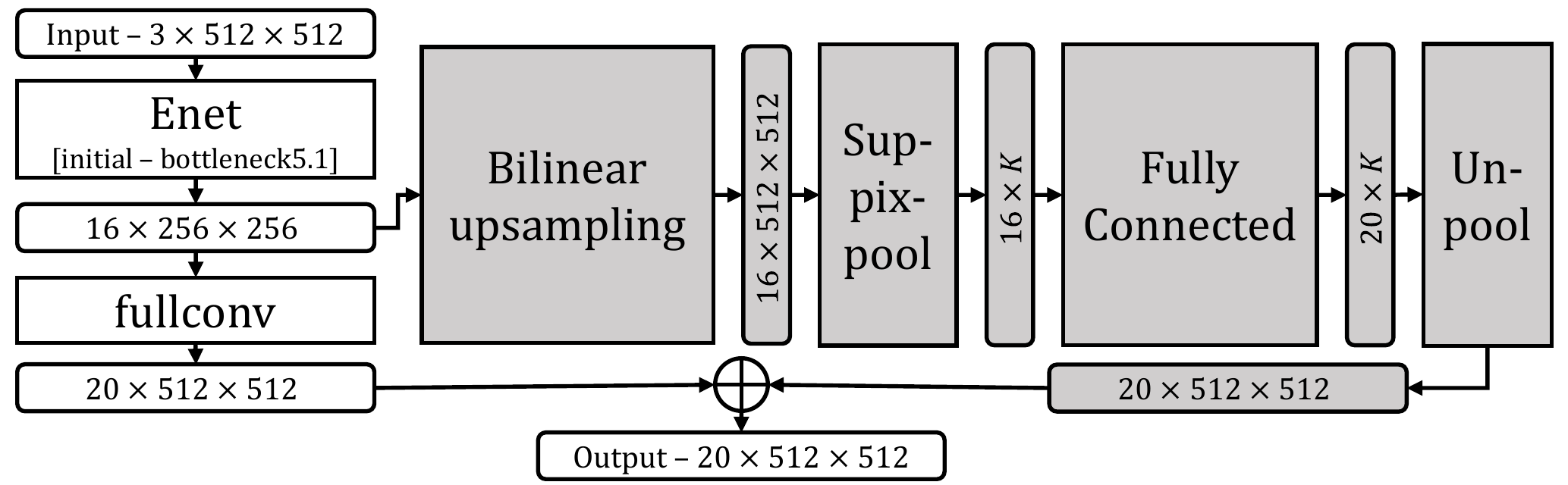}
\caption{Integration of the superpixel pooling layer in the ENet~\cite{paszke2016enet} architecture. The gray blocks are added to the original network.}
\label{fig:ENetDesign}
\end{figure}
We demonstrate the relevance of adding superpixel pooling layer to ENet~\cite{paszke2016enet}, a semantic segmentation architecture tuned for speed. We show that the superpixel pooling layer can greatly enhance the quality of resulting segmentations, for a minimal inference overhead.

Similarly to the V3 set-up for the \textit{VoxResNet} architecture, we modify the \textit{ENet} architecture by adding a parallel superpixel branch to the end of the network. We again pool the activations in the feature map over the superpixels. 
Then, we add a fully-connected layer to map the superpixel-wise activations to the output space. Finally, the results of the two branches are combined in a pixel-wise sum. 
A schematic representation of the design can be found in~\Cref{fig:ENetDesign}. 
We compute superpixel segmentations on GPU during image preprocessing with gSLICr~\cite{ren2015gslicr}, which introduces minimal overhead. 
In our experiments, we load the parameters of the original authors and finetune the network on the Cityscapes dataset~\cite{Cordts2016Cityscapes}. 

\subsubsection{Results}
We finetuned the network using SLIC superpixels as well as rectangular superpixels with the number of superpixels $K$ among 500, 1000, 5000 and 10 000. Based on the validation data we found that the best results are obtained with $K=1000$.~\Cref{tab:ENet} gives a comparison between the accuracy of this network and the original ENet architecture evaluated on the Cityscapes~\cite{Cordts2016Cityscapes} test data. 
\begin{table}[h]
\centering
\caption{Results on the Cityscapes test data (\%).}
\label{tab:ENet}
\begin{tabular}{@{}lcccc@{}}
\toprule
                   & IoU Classes & iIoU Classes & IoU Categories & iIoU Categories \\ \midrule
ENet               & 58.3        & 34.4         & 80.3          & 64.0           \\
ENet + superpixels &\textbf{61.3}        & \textbf{36.8}         & \textbf{82.1}           & \textbf{66.9}            \\ \bottomrule
\end{tabular}
\end{table}

\Cref{tab:ENet_cat} shows the IoU scores on the different object categories defined in Cityscapes (as supersets of the object classes). While the superpixels bring an improvement in all categories, we observe that the effect is largest on objects with finer details and distinct edges that appear strongly against the background, as in the \texttt{objects} and \texttt{humans} categories. 
This is confirmed by the class-level scores: the scores on \texttt{traffic light}, \texttt{traffic sign} and \texttt{rider} increase by more than 8\%, while the improvement on other classes are between 0.2\% and 4\%. 
Since superpixels tend to well preserve highly-contrasted edges, they serve as an effective prior for these classes and allow the network to recover the edges with finer details. 
\begin{table}[ht] \small
\centering
\caption{Category-level results on the Cityscapes test data (\%).}
\label{tab:ENet_cat}
\begin{tabular}{@{}lccccccc@{}}
\toprule
                 & flat          & nature        & object        & sky           & construction  & human         & vehicle       \\ \midrule
Enet             & 97.3          & 88.3          & 46.8          & 90.6          & 85.4          & 65.5          & 88.9          \\
Enet + superpixels & \textbf{97.7} & \textbf{89.2} & \textbf{50.9} & \textbf{92.5} & \textbf{86.6} & \textbf{68.3} & \textbf{89.6} \\ \bottomrule
\end{tabular}
\end{table}

Using an Nvidia Titan X (Pascal) GPU, the inference time on images of resolution $1024 \times 512$ (used in our tests on the cityscapes dataset) is \SI{13}{\ms} 
for ENet with superpixels. By comparison, it is \SI{11}{\ms} seconds for the base network. 
We believe that further optimizations to the superpixel pooling layer implementation on GPU are possible and would further decrease the overhead induced by the superpixel pooling layer.

\begin{figure}[ht]
\subfigure[ENet output]{\includegraphics[width=0.33\linewidth]{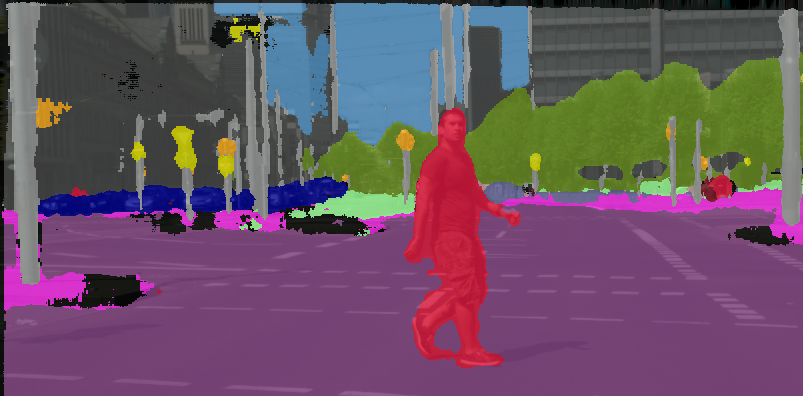}%
}\hfill%
\subfigure[Ground truth]{\includegraphics[width=0.33\linewidth]{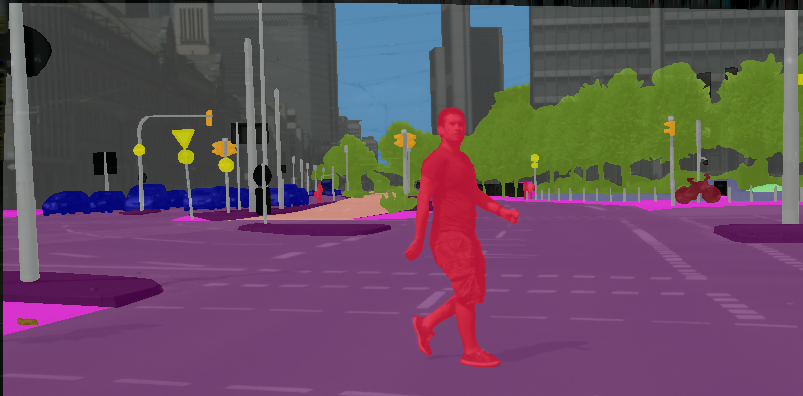}%
}\hfill%
\subfigure[ENet + superpixels]{\includegraphics[width=0.33\linewidth]{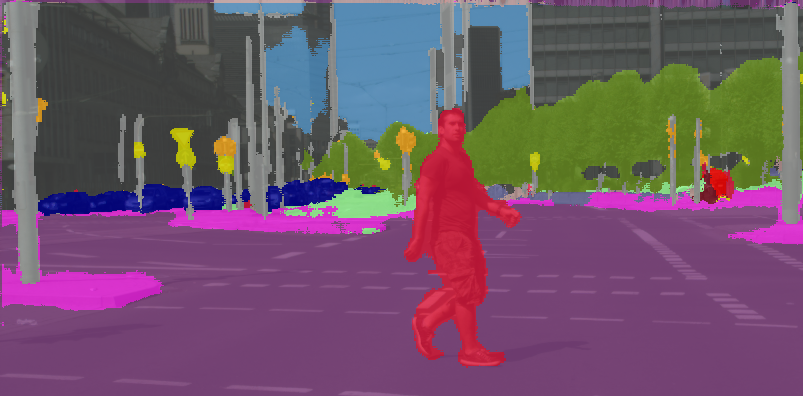}%
}\caption{ENet outputs (details) without and with a superpixel pooling branch.}
\label{fig:cityscapesViz}
\end{figure}

Figure~\ref{fig:cityscapesViz} shows a visual comparison of an output segmentation of ENet, without and with the superpixel pooling branch.
It indicates closer adherence to hard edges (note for example the feet of the pedestrian), which was already indicated by the category-wise results in~\Cref{tab:ENet_cat}. Furthermore, we observe that classes that tend to represent large homogeneous areas, like \texttt{road} or \texttt{pavement} benefit a lot from intermediate-level feature pooling introduced by the superpixel pooling layer. This is evidenced by the spurious undefined regions on the left for example, which are removed by the grouping of these regions into superpixels. The increased accuracy both on hard, precise edges that make up fine details as well as the increased homogeneity of larger regions result in the increased test accuracies shown in Tables~\ref{tab:ENet}~--~\ref{tab:ENet_cat}.

\section{Conclusion}
In this paper, we propose a simple but efficient GPU-implementation of a superpixel (or supervoxel) pooling layer for CNNs and provide experimental results of this layer within two existing segmentation networks: \textit{VoxResNet}~\cite{chen2016voxresnet} and \textit{Enet}~\cite{paszke2016enet}.
We show empirically that the segmentation accuracy of the network consistently increases when the superpixel pooling layer is used in conjunction with pixel-wise segmentation. This allows the network to use contextual information from the superpixels, but make corrections on the pixel level.
Furthermore, we observe that the benefits of using a superpixel pooling layer tend to increase with simpler networks. 
This motivates the use of superpixel pooling for real-time segmentation applications.
Source code is available at \projecturl.

\paragraph{Acknowledgements}

This work is partially funded by
Internal Funds KU Leuven and an Amazon Research Award. This work made use of a hardware donation from the Facebook GPU Partnership program. We
acknowledge support from the Research Foundation - Flanders
(FWO) through project number G0A2716N.

\bibliography{egbib}

\begin{thebibliography}{10}

\bibitem{achanta2012slic}
Radhakrishna Achanta, Appu Shaji, Kevin Smith, Aurelien Lucchi, Pascal Fua, and
  Sabine S{\"u}sstrunk.
\newblock {SLIC} superpixels compared to state-of-the-art superpixel methods.
\newblock {\em IEEE Transactions on Pattern Analysis and Machine Intelligence},
  34(11):2274--2282, 2012.

\bibitem{chen2016voxresnet}
Hao Chen, Qi~Dou, Lequan Yu, Jing Qin, and Pheng-Ann Heng.
\newblock {VoxResNet}: Deep voxelwise residual networks for brain segmentation
  from {3D} {MR} images.
\newblock {\em NeuroImage}, 170:446--455, 2018.

\bibitem{Cordts2016Cityscapes}
Marius Cordts, Mohamed Omran, Sebastian Ramos, Timo Rehfeld, Markus Enzweiler,
  Rodrigo Benenson, Uwe Franke, Stefan Roth, and Bernt Schiele.
\newblock The {Cityscapes} dataset for semantic urban scene understanding.
\newblock In {\em Proc.\ of the IEEE Conference on Computer Vision and Pattern
  Recognition (CVPR)}, 2016.

\bibitem{he2016deep}
Kaiming He, Xiangyu Zhang, Shaoqing Ren, and Jian Sun.
\newblock Deep residual learning for image recognition.
\newblock In {\em Proceedings of the IEEE Conference on Computer Vision and
  Pattern Recognition}, pages 770--778, 2016.

\bibitem{scipy}
Eric Jones, Travis Oliphant, Pearu Peterson, et~al.
\newblock {SciPy}: Open source scientific tools for {Python}, 2001--.
\newblock [Online; accessed 10/04/2018].

\bibitem{kwak2017weakly}
Suha Kwak, Seunghoon Hong, Bohyung Han, et~al.
\newblock Weakly supervised semantic segmentation using superpixel pooling
  network.
\newblock In {\em AAAI}, pages 4111--4117, 2017.

\bibitem{numba}
Siu~Kwan Lam, Antoine Pitrou, and Stanley Seibert.
\newblock Numba: A {LLVM}-based {Python} {JIT} compiler.
\newblock In {\em Proceedings of the Second Workshop on the LLVM Compiler
  Infrastructure in HPC}, pages 7:1--7:6, 2015.

\bibitem{paszke2016enet}
Adam Paszke, Abhishek Chaurasia, Sangpil Kim, and Eugenio Culurciello.
\newblock {ENet}: A deep neural network architecture for real-time semantic
  segmentation.
\newblock {\em arXiv preprint arXiv:1606.02147}, 2016.

\bibitem{ren2015gslicr}
Carl~Yuheng Ren, Victor~Adrian Prisacariu, and Ian~D Reid.
\newblock {gSLICr}: {SLIC} superpixels at over 250{Hz}.
\newblock {\em arXiv preprint arXiv:1509.04232}, 2015.

\bibitem{rohlfing2012image}
Torsten Rohlfing.
\newblock Image similarity and tissue overlaps as surrogates for image
  registration accuracy: widely used but unreliable.
\newblock {\em IEEE Transactions on Medical Imaging}, 31(2):153--163, 2012.

\bibitem{stutz2017superpixels}
David Stutz, Alexander Hermans, and Bastian Leibe.
\newblock Superpixels: An evaluation of the state-of-the-art.
\newblock {\em Computer Vision and Image Understanding}, 166:1--27, 2018.

\bibitem{van2012seeds}
Michael Van~den Bergh, Xavier Boix, Gemma Roig, Benjamin de~Capitani, and Luc
  Van~Gool.
\newblock {SEEDS}: Superpixels extracted via energy-driven sampling.
\newblock In Andrew Fitzgibbon, Svetlana Lazebnik, Pietro Perona, Yoichi Sato,
  and Cordelia Schmid, editors, {\em Computer Vision -- ECCV 2012}, pages
  13--26. Springer, 2012.

\bibitem{yang2017improved}
Yadong Yang and Xiaofeng Wang.
\newblock Improved {CNN} based on super-pixel segmentation.
\newblock In {\em International Conference on Intelligence Science}, pages
  305--310. Springer, 2017.

\bibitem{yao2015real}
Jian Yao, Marko Boben, Sanja Fidler, and Raquel Urtasun.
\newblock Real-time coarse-to-fine topologically preserving segmentation.
\newblock In {\em Proceedings of the IEEE Conference on Computer Vision and
  Pattern Recognition}, pages 2947--2955, 2015.

\end{thebibliography}

\end{document}